\documentclass[11pt]{article}
\usepackage{acl2016}
\usepackage{times}
\usepackage{url}
\usepackage{latexsym}

\usepackage[utf8]{inputenc}
\usepackage{amssymb}
\usepackage{amsmath}
\usepackage{verbatim}
\usepackage{graphicx}
\usepackage{hhline}

\usepackage{tikz}
\usetikzlibrary{positioning,arrows,patterns}
\tikzset{execute at begin node=\strut}
\definecolor{lila}{RGB}{159,114,207}
\definecolor{orange}{RGB}{230,130,50}
\colorlet{plila}{lila!60!white}
\colorlet{porange}{orange!60!white}
\definecolor{grau}{RGB}{192,192,192}

\newcommand{\dth}{\frac{\partial}{\partial \theta}}
\newcommand{\dthb}[1]{\dth\left(#1\right)}
\newcommand{\ene}[1]{\exp\left(-E^b(#1)\right)}
\newcommand{\xp}{{x^\prime}}
\newcommand{\spr}{{s^\prime}}
\newcommand{\cp}{{c^\prime}}
\newcommand{\E}[2]{\mathbb{E}_{#1}\left[#2\right]}

\aclfinalcopy 

\title{Functional Distributional Semantics}

\author{Guy Emerson and Ann Copestake \\
  Computer Laboratory \\
  University of Cambridge \\
  {\tt \{gete2,aac10\}@cam.ac.uk} \\}

\date{}

\begin{document}
\maketitle

\begin{abstract}
Vector space models have become popular in distributional semantics,
despite the challenges they face in capturing various semantic phenomena.
We propose a novel probabilistic framework which draws on both formal semantics
and recent advances in machine learning.
In particular, we separate predicates from the entities they refer to,
allowing us to perform Bayesian inference based on logical forms.
We describe an implementation of this framework
using a combination of Restricted Boltzmann Machines and feedforward neural networks.
Finally, we demonstrate the feasibility of this approach
by training it on a parsed corpus
and evaluating it on established similarity datasets.

\end{abstract}

\section{Introduction}

Current approaches to distributional semantics generally involve
representing words as points in a high-dimensional vector space.
However, vectors do not provide `natural' composition operations
that have clear analogues with operations in formal semantics,
which makes it challenging to perform inference,
or capture various aspects of meaning studied by semanticists.
This is true whether the vectors are constructed using a \emph{count} approach
(e.g.\ Turney and Pantel, 2010) \nocite{turney2010vector}
or an \emph{embedding} approach
(e.g.\ Mikolov et al., 2013), \nocite{mikolov2013vector}
and indeed \newcite{levy2014embed} showed that there are close links between them.
Even the tensorial approach described by
\newcite{coecke2010tensor} and \newcite{baroni2014tensor},
which naturally captures argument structure,
does not allow an obvious account of context dependence, or logical inference.

In this paper, we build on insights drawn from formal semantics,
and seek to learn representations which have a more natural logical structure,
and which can be more easily integrated with other sources of information.

Our contributions in this paper are
to introduce a novel framework for distributional semantics,
and to describe an implementation and training regime in this framework.
We present some initial results to demonstrate that training this model is feasible.

\section{Formal Framework of Functional Distributional Semantics}

In this section, we describe our framework,
explaining the connections to formal semantics,
and defining our probabilistic model.
We first motivate representing predicates with functions,
and then explain how these functions can be incorporated into
a representation for a full utterance.

\subsection{Semantic Functions}
\label{sec:sem-func}

\begin{figure*}
\centering
\vspace*{-2mm}

\begin{tikzpicture}[on grid, node distance = 2cm, x = 2cm, y = 2cm]
\tikzstyle{hash}=[pattern=north east lines]

\node (green) [anchor=north, yshift=-1mm] {\includegraphics[width=.08\textwidth]{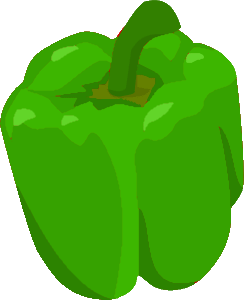}};
\node (yellow) [right = of green] {\includegraphics[width=.08\textwidth]{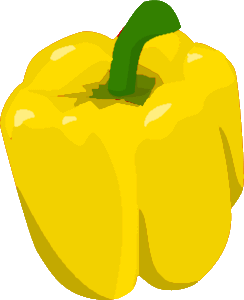}};
\node (red) [right = of yellow] {\includegraphics[width=.08\textwidth]{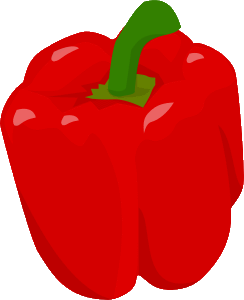}};
\node (purple) [right = of red] {\includegraphics[width=.08\textwidth]{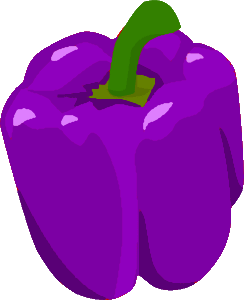}};
\node (blue) [right = of purple] {\includegraphics[width=.08\textwidth]{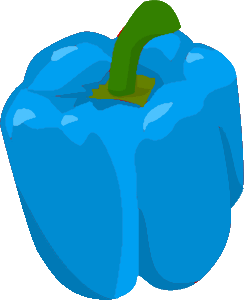}};
\node (carrot) [right = of blue] {\includegraphics[width=.08\textwidth]{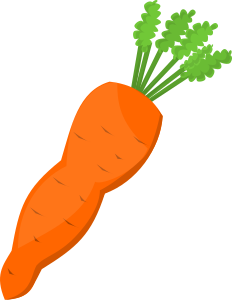}};
\node (cucumber) [right = of carrot] {\includegraphics[width=.08\textwidth]{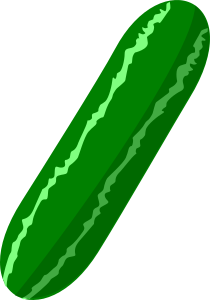}};

\fill (-0.2, 0) rectangle +(0.16, 1);
\fill (0.8, 0) rectangle +(0.16, 1);
\fill (1.8, 0) rectangle +(0.16, 1);
\fill (2.8, 0) rectangle +(0.16, 0.8);
\fill (3.8, 0) rectangle +(0.16, 0.75);

\fill[hash] (0.04, 0) rectangle +(0.16, 0.3);
\fill[hash] (1.04, 0) rectangle +(0.16, 0.3);
\fill[hash] (2.04, 0) rectangle +(0.16, 0.35);
\fill[hash] (3.04, 0) rectangle +(0.16, 0.05);

\fill (-0.5, -0.01) rectangle +(0.02, 1.02);
\fill (-0.5, 0.01) rectangle +(-0.05, -0.02) node[left] {0};
\fill (-0.5, 1.01) rectangle +(-0.05, -0.02) node[left] {1};

\end{tikzpicture}

\vspace*{-4mm}

\caption{Comparison between a \emph{semantic function} and a \emph{distribution} over a space of entities.
The vegetables depicted above
(five differently coloured bell peppers, a carrot, and a cucumber)
form a discrete semantic space $\mathcal{X}$.
We are interested in the truth $t$ of the predicate for \emph{bell pepper} for an entity $x \in \mathcal{X}$. \\
Solid bars: the semantic function $P(t|x)$
represents how much each entity is considered to be a pepper,
and is bounded between $0$~and~$1$;
it is high for all the peppers, but slightly lower for atypical colours. \\
Shaded bars: the distribution $P(x|t)$
represents our belief about an entity if all we know is that the predicate for \emph{bell pepper} applies to it;
the probability mass must sum to $1$, so it is split between the peppers,
skewed towards typical colours, and excluding colours believed to be impossible.}
\vspace*{-4mm}

\label{fig:pepper}
\end{figure*}
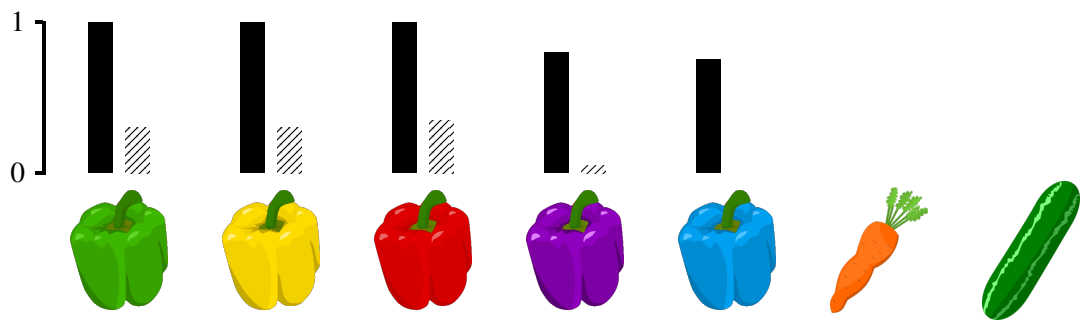

We begin by assuming an extensional model structure,
as standard in formal semantics
\cite{kamp1993discourse,cann1993semantics,allan2001semantics}.
In the simplest case, a model contains a set of entities,
which predicates can be true or false of.
Models can be endowed with additional structure,
such as for plurals \cite{link2002plural},
although we will not discuss such details here.
For now, the important point is that we should
separate the representation of a predicate
from the representations of the entities it is true of.

We generalise this formalisation of predicates by
treating truth values as random variables,\footnote{
  The move to replace absolute truth values with probabilities
  has parallels in much computational work based on formal logic.
  For example, \newcite{garrette2011markov}
  incorporate distributional information in a Markov Logic Network \cite{richardson2006markov}.
  However, while their approach allows probabilistic inference,
  they rely on existing distributional vectors,
  and convert similarity scores to weighted logical formulae.
  Instead, we aim to learn representations
  which are directly interpretable within in a probabilistic logic.}
which enables us to apply Bayesian inference.
For any entity, we can ask which predicates are true of it (or `applicable' to it).
More formally, if we take entities to lie in some semantic space $\mathcal{X}$
(whose dimensions may denote different features),
then we can take the meaning of a predicate to be
a function from $\mathcal{X}$ to values in the interval $[0,1]$,
denoting how likely a speaker is
to judge the predicate applicable to the entity.
This judgement is variable between speakers \cite{labov1973cup},
and for borderline cases,
it is even variable for one speaker at different times \cite{mccloskey1978judge}.

Representing predicates as functions
allows us to naturally capture vagueness
(a predicate can be equally applicable to multiple points),
and using values between 0 and 1
allows us to naturally capture gradedness
(a predicate can be more applicable to some points than to others).
To use Labov's example, the predicate for \emph{cup}
is equally applicable to vessels of different shapes and materials,
but becomes steadily less applicable to wider vessels.

We can also view such a function as a classifier --
for example, the semantic function for the predicate for \emph{cat}
would be a classifier separating cats from non-cats.
This ties in with a view of concepts as abilities, as proposed in both
philosophy \cite{dummett1978know,kenny2010concept},
and cognitive science \cite{murphy2002concept,bennett2008neuro}.
A similar approach is taken by \newcite{larsson2013percept},
who argues in favour of representing perceptual concepts as classifiers of perceptual input.

Note that these functions do not directly define
probability distributions over entities.
Rather, they define binary-valued \emph{conditional} distributions,
\emph{given an entity}.
We can write this as $P(t|x)$,
where $x$ is an entity,
and $t$ is a stochastic truth value.
It is only possible to get a corresponding
distribution over entities given a truth value, $P(x|t)$,
if we have some background distribution $P(x)$.
If we do, we can apply Bayes' Rule to get ${P(x|t)\propto P(t|x)P(x)}$.
In other words, the truth of an expression
depends crucially on our knowledge of the situation.
This fits neatly within a verificationist view of truth,
as proposed by \newcite{dummett1976meaning},
who argues that to understand a sentence
is to know how we could verify or falsify it.

By using both $P(t|x)$ and $P(x|t)$,
we can distinguish between underspecification and uncertainty as two kinds of `vagueness'.
In the first case, we want to state partial information about an entity,
but leave other features unspecified;
$P(t|x)$ represents which kinds of entity could be described by the predicate,
regardless of how likely we think the entities are.
In the second case, we have uncertain knowledge about the entity;
$P(x|t)$ represents which kinds of entity we think are likely for this predicate,
given all our world knowledge.

For example, bell peppers come in many colours,
most typically green, yellow, orange or red.
As all these colours are typical, the semantic function for the predicate for \textit{bell pepper}
would take a high value for each.
In contrast, to define a probability distribution over entities,
we must split probability mass between different colours,\footnote{
  In fact, colour would be most properly treated as a continuous feature.
  In this case, $P\left(x\right)$ must be a probability density function, not a probability mass function,
  whose value would further depend on the parametrisation of the space.}
and for a large number of colours,
we would only have a small probability for each.
As purple and blue are atypical colours for a pepper,
a speaker might be less willing to label such a vegetable a pepper,
but not completely unwilling to do so --
this linguistic knowledge belongs to the semantic function for the predicate.
In contrast, after observing a large number of peppers,
we might conclude that blue peppers do not exist,
purple peppers are rare,
green peppers common,
and red peppers more common still --
this world knowledge belongs to the probability distribution over entities.
The contrast between these two quantities
is depicted in figure~\ref{fig:pepper},
for a simple discrete space.

\subsection{Incorporation with Dependency Minimal Recursion Semantics}
\label{sec:dmrs}

Semantic dependency graphs have become popular in NLP.
We use Dependency Minimal Recursion Semantics (DMRS) \cite{copestake2005mrs,copestake2009dmrs},
which represents meaning
as a directed acyclic graph:
nodes represent predicates/entities
(relying on a one-to-one correspondence between them)
and links (edges) represent argument structure and scopal constraints.
Note that we assume a neo-Davidsonian approach
\cite{davidson1967event,parsons1990event},
where events are also treated as entities,
which allows a better account of adverbials, among other phenomena.

For example (simplifying a little), to represent \emph{``the dog barked''},
we have three nodes, for the predicates \emph{the}, \emph{dog}, and \emph{bark},
and two links: an \textsc{arg1} link from \emph{bark} to \emph{dog},
and a \textsc{rstr} link from \emph{the} to \emph{dog}.
Unlike syntactic dependencies,
DMRS abstracts over semantically equivalent expressions,
such as \emph{``dogs chase cats''} and \emph{``cats are chased by dogs''}.
Furthermore, unlike other types of semantic dependencies,
including Abstract Meaning Representations \cite{banarescu2012amr},
and Prague Dependencies \cite{bohmova2003prague},
DMRS is interconvertible with MRS,
which can be given a direct logical interpretation.

We deal here with the extensional fragment of language,
and while we can account for different quantifiers in our framework,
we do not have space to discuss this here --
for the rest of this paper, we neglect quantifiers,
and the reader may assume that all variables are existentially quantified.

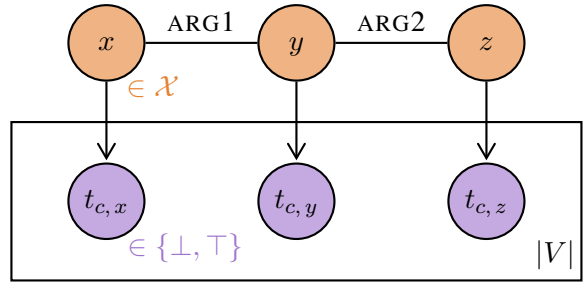
\begin{figure}
\centering

\begin{tikzpicture}[on grid, node distance=21mm and 25mm, x=25mm, y=21mm, style=thick, inner sep=0pt]
\tikzstyle{ent}=[circle, draw, minimum size=10mm, fill=porange]
\tikzstyle{val}=[circle, draw, minimum size=10mm, fill=plila]
\tikzstyle{tok}=[circle, draw, minimum size=10mm, fill=grau]


\node[ent] (y) {$y$} ;
\node[ent, right=of y] (z) {$z$} ;
\node[ent, left=of y] (x) {$x$} ;

\draw (y) -- (z) node[midway, above, color=black] {\textsc{arg2}} ;
\draw (y) -- (x) node[midway, above, color=black] {\textsc{arg1}} ;

\node[below right=3.5mm and 2.5mm of x, anchor=north west] {\textcolor{orange}{$\in \mathcal{X}$}} ;


\draw (-1.5,-0.5) rectangle (1.5,-1.5) ;

\node[val, below=of x] (tx) {$t_{c,\,x}$} ;
\node[val, below=of y] (ty) {$t_{c,\,y}$} ;
\node[val, below=of z] (tz) {$t_{c,\,z}$} ;

\tikzset{style=ultra thick}
\draw[-angle 60] (x) -- (tx);
\draw[-angle 60] (y) -- (ty);
\draw[-angle 60] (z) -- (tz);

\node[below right=3.5mm and 2.5mm of tx, anchor=north west] {\textcolor{lila}{$\in \left\{\bot,\top\right\}$}} ;
\node[xshift=-2ex, yshift=2ex] at (1.5, -1.5) {$|V|$};

\end{tikzpicture}

\caption{A situation composed of three entities.
Top row: the entities $x$, $y$, and $z$ lie in a semantic space~$\mathcal{X}$,
jointly distributed according to DMRS links.
Bottom row: each predicate $c$ in the vocabulary $V$
has a stochastic truth value for each entity.
}
\label{fig:graph-model}
\vspace*{-3mm}

\end{figure}

We can use the structure of a DMRS graph
to define a probabilistic graphical model.
This gives us a distribution over lexicalisations of the graph --
given an abstract graph structure,
where links are labelled but nodes are not,
we have a process to generate a predicate for each node.
Although this process is different for each graph structure,
we can share parameters between them
(e.g.\ according to the labels on links).
Furthermore, if we have a distribution over graph structures,
we can incorporate that in our generative process,
to produce a distribution over lexicalised graphs.

The entity nodes can be viewed as together representing a situation,
in the sense of \newcite{barwise1983situation}.
We want to be able to represent the entities
without reference to the predicates --
intuitively, the world is the same however we choose to describe it.
To avoid postulating causal structure amongst the entities
(which would be difficult for a large graph),
we can model the entity nodes as an undirected graphical model,
with edges according to the DMRS links.
The edges are undirected in the sense that they don't impose conditional dependencies.
However, this is still compatible with having `directed' semantic dependencies --
the probability distributions are not symmetric, which maintains the asymmetry of DMRS links.

Each node takes values in the semantic space $\mathcal{X}$,
and the network defines a joint distribution over entities,
which represents our knowledge
about which situations are likely or unlikely.
An example is shown in the top row of figure~\ref{fig:graph-model},
of an entity $y$ along with its two arguments $x$ and $z$ --
these might represent an event,
along with the agent and patient involved in the event.
The structure of the graph means that we can factorise
the joint distribution $P\left(x,y,z\right)$ over the entities
as being proportional to the product $P\left(x,y\right) P\left(y,z\right)$.

For any entity, we can ask which predicates are true of it.
We can therefore introduce a node for every predicate in the vocabulary,
where the value of the node is either true ($\top$) or false ($\bot$).
Each of these predicate nodes has a single directed link from the entity node,
with the probability of the node being true being determined by the predicate's semantic function,
i.e.\ ${P\left(t_{c,\,x}=\top\middle|x\right)=t_c(x)}$.
This is shown in the second row of figure~\ref{fig:graph-model},
where the plate denotes that these nodes are repeated for each predicate $c$ in the vocabulary $V$.
For example, if the situation represented a dog chasing a cat,
then nodes like $t_{dog,\,x}$, $t_{animal,\,x}$, and $t_{pursue,\,y}$
would be true (with high probability),
while $t_{democracy,\,x}$ or $t_{dog,\,z}$ would be false (with high probability).

\begin{figure}
\centering

\begin{tikzpicture}[on grid, node distance=21mm and 25mm, x=25mm, y=21mm, style=thick, inner sep=0pt]
\tikzstyle{ent}=[circle, draw, minimum size=10mm, fill=porange]
\tikzstyle{val}=[circle, draw, minimum size=10mm, fill=plila]
\tikzstyle{tok}=[circle, draw, minimum size=10mm, fill=grau]


\node[ent] (y) {$y$} ;
\node[ent, right=of y] (z) {$z$} ;
\node[ent, left=of y] (x) {$x$} ;

\draw (y) -- (z) node[midway, above, color=black] {\textsc{arg2}} ;
\draw (y) -- (x) node[midway, above, color=black] {\textsc{arg1}} ;

\node[below right=3.5mm and 2.5mm of x, anchor=north west] {\textcolor{orange}{$\in \mathcal{X}$}} ;


\draw (-1.5,-0.5) rectangle (1.5,-1.5) ;

\node[val, below=of x] (tx) {$t_{c,\,x}$} ;
\node[val, below=of y] (ty) {$t_{c,\,y}$} ;
\node[val, below=of z] (tz) {$t_{c,\,z}$} ;

\tikzset{style=ultra thick}
\draw[-angle 60] (x) -- (tx);
\draw[-angle 60] (y) -- (ty);
\draw[-angle 60] (z) -- (tz);

\node[below right=3.5mm and 2.5mm of tx, anchor=north west] {\textcolor{lila}{$\in \left\{\bot,\top\right\}$}} ;
\node[xshift=-2ex, yshift=2ex] at (1.5, -1.5) {$|V|$};


\node[tok, below=of tx] (a) {$p$} ;
\node[tok, below=of ty] (b) {$q$} ;
\node[tok, below=of tz] (c) {$r$} ;

\draw[-angle 60] (tx) -- (a);
\draw[-angle 60] (ty) -- (b);
\draw[-angle 60] (tz) -- (c);

\node[below right=3.5mm and 2.5mm of a, anchor=north west] {$\in V$} ;

\end{tikzpicture}
\vspace*{-3mm}

\caption{The probabilistic model in figure~\ref{fig:graph-model},
extended to generate utterances.
Each predicate in the bottom row
is chosen out of all predicates which are true for the corresponding entity.
}
\label{fig:graph-token}
\vspace*{-3mm}

\end{figure}
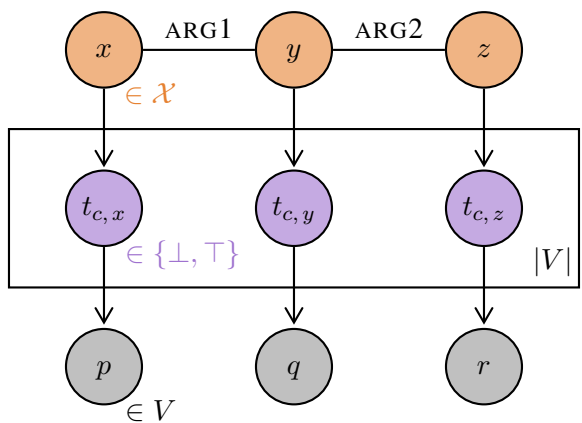

The probabilistic model described above closely matches traditional model-theoretic semantics.
However, while we could stop our semantic description there,
we do not generally observe truth-value judgements for all predicates at once;\footnote{
  This corresponds to what \newcite{copestake2012lexical} call an \emph{ideal distribution}.
  If we have access to such information, we only need the two rows given in figure~\ref{fig:graph-model}.}
rather, we observe utterances, which have specific predicates.
We can therefore define a final node for each entity,
which takes values over predicates in the vocabulary,
and which is conditionally dependent on the truth values of all predicates.
This is shown in the bottom row of figure~\ref{fig:graph-token}.
Including these final nodes
means that we can train such a model on observed utterances.
The process of choosing a predicate from the true ones may be complex,
potentially depending on speaker intention and other pragmatic factors --
but in section~\ref{sec:impl}, we will simply choose a true predicate at random (weighted by frequency).

The separation of entities and predicates
allows us to naturally capture context-dependent meanings.
Following the terminology of \newcite{quine1960meaning},
we can distinguish context-independent \emph{standing} meaning
from context-dependent \emph{occasion} meaning.
Each predicate type has a corresponding semantic function
-- this represents its standing meaning.
Meanwhile, each predicate token has a corresponding entity,
for which there is a posterior distribution over the semantic space,
conditioning on the rest of the graph and any pragmatic factors
-- this represents its occasion meaning.

Unlike previous approaches to context dependence, such as
\newcite{dinu2012compare}, \newcite{erk2008context}, and \newcite{thater2011context},
we represent meanings in and out of context by different kinds of object,
reflecting a type/token distinction.
Even \newcite{herbelot2015name},
who explicitly contrasts individuals and kinds,
embeds both in the same space.

As an example of how this separation of predicates and entities can be helpful,
suppose we would like
\emph{``dogs chase cats''} and \emph{``cats chase mice''}
to be true in a model, but
\emph{``dogs chase mice''} and \emph{``cats chase cats''}
to be false.
In other words, there is a dependence between the verb's arguments.
If we represent each predicate by a single vector,
it is not clear how to capture this.
However, by separating predicates from entities,
we can have two different entities which \emph{chase} is true of,
where one co-occurs with a dog-entity \textsc{arg1} and cat-entity \textsc{arg2},
while the other co-occurs with a cat-entity \textsc{arg1} and a mouse-entity \textsc{arg2}.

\section{Implementation}
\label{sec:impl}

In the previous section, we described a general framework for probabilistic semantics.
Here we give details of one way that such a framework can be implemented for distributional semantics,
keeping the architecture as simple as possible.

\subsection{Network Architecture}

We take the semantic space $\mathcal{X}$ to be a set of binary-valued vectors,\footnote{
  We use the term \textit{vector} in the computer science sense of a linear array,
  rather than in the mathematical sense of a point in a vector space.}
$\{0,1\}^N$.
A situation~$s$ is then composed of entity vectors ${x^{(1)},\cdots,x^{(K)}\in \mathcal{X}}$
(where the number of entities $K$ may vary),
along with links between the entities.
We denote a link from $x^{(n)}$ to $x^{(m)}$ with label~$l$
as: ${x^{(n)}\xrightarrow{l} x^{(m)}}$.
We define the background distribution over situations
using a Restricted Boltzmann Machine (RBM)
\cite{smolensky1986rbm,hinton2006rbm},  
but rather than having connections between hidden and visible units,
we have connections between components of entities,
according to the links.

The probability of the network being in the particular configuration $s$
depends on the \emph{energy} of the configuration, $E^b(s)$,
as shown in equations~(\ref{eqn:rbm})-(\ref{eqn:rbm-z}).
A high energy denotes an unlikely configuration.
The energy depends on the edges of the graphical model, plus bias terms,
as shown in~(\ref{eqn:link}).
Note that we follow the Einstein summation convention,
where repeated indices indicate summation;
although this notation is not typical in NLP,
we find it much clearer than matrix-vector notation,
particularly for higher-order tensors.
Each link label $l$ has a corresponding weight matrix $W^{(l)}$,
which determines the strength of association
between components of the linked entities.
The first term in~(\ref{eqn:link})
sums these contributions over all links
${x^{(n)}\xrightarrow{l} x^{(m)}}$
between entities.
We also introduce bias terms,
to control how likely an entity vector is, independent of links.
The second term in~(\ref{eqn:link}) sums the biases over all entities $x^{(n)}$.
\begin{equation}
P(s) = \frac{1}{Z}\ene{s}
\label{eqn:rbm}
\end{equation}
\begin{equation}
Z = \sum_\spr\ene{\spr}
\label{eqn:rbm-z}
\end{equation}
\begin{equation}
-E^b(s) = \!\!\!\sum_{x^{(n)}\xrightarrow{l} x^{(m)}\!\!}\!\!\!\!\!
  W^{(l)}_{ij}x^{(n)}_i x^{(m)}_j
  - \sum_{x^{(n)}} b^{\ }_i x^{(n)}_i \!
\label{eqn:link}
\end{equation}

Furthermore, since sparse representations
have been shown to be beneficial in NLP,
both for applications and for interpretability of features
\cite{murphy2012sparse,faruqui2015sparse},
we can enforce sparsity in these entity vectors
by fixing a specific number of units to be active at any time.
\newcite{swersky2012cardinality} introduce this RBM variant as the Cardinality RBM,
and also give an efficient exact sampling procedure using belief propagation.
Since we are using sparse representations,
we also assume that all link weights are non-negative.

Now that we've defined the background distribution over situations,
we turn to the semantic functions $t_c$, which map entities $x$ to probabilities.
We implement these as feedforward networks,
as shown in~(\ref{eqn:semf})-(\ref{eqn:semf-z}).
For simplicity, we do not introduce any hidden layers.
Each predicate $c$ has a vector of weights $W^{\prime(c)}$,
which determines the strength of association
with each dimension of the semantic space,
as well as a bias term $b^{\prime(c)}$.
These together define the energy $E^p_{\vphantom{!}}(x,c)$
of an entity $x$ with the predicate,
which is passed through a sigmoid function
to give a value in the range $[0,1]$.
\begin{equation}
t_c(x) = \sigma(-E^p(x,c)) = \frac{1}{1+\exp\left({E^p_{\vphantom{1}}}\right)}
\label{eqn:semf}
\end{equation}
\begin{equation}
-E^p(x,c) = W^{\prime(c)}_i x^{\ }_i - b^{\prime(c)}
\label{eqn:semf-z}
\end{equation}

Given the semantic functions,
choosing a predicate for a entity can be hard-coded, for simplicity.
The probability of choosing a predicate $c$ for an entity $x$
is weighted by the predicate's frequency $f_c$
and the value of its semantic function $t_c(x)$
(how true the predicate is of the entity),
as shown in~(\ref{eqn:choose})-(\ref{eqn:choose-z}).
This is a mean field approximation to the stochastic truth values shown in figure~\ref{fig:graph-token}.
\begin{equation}
P(c|x) = \frac{1}{Z_x}f_c t_c(x) 
\label{eqn:choose}
\end{equation}
\begin{equation}
Z_x = \sum_\cp f_\cp t_\cp(x)
\label{eqn:choose-z}
\end{equation}

\subsection{Learning Algorithm}

To train this model,
we aim to maximise the likelihood of observing the training data --
in Bayesian terminology, this is \emph{maximum a posteriori} estimation.
As described in section~\ref{sec:dmrs},
each data point is a lexicalised DMRS graph,
while our model defines distributions over lexicalisations of graphs.
In other words, we take as given the observed distribution over abstract graph structures
(where links are given, but nodes are unlabelled),
and try to optimise how the model generates predicates
(via the parameters $W^{(l)}_{ij}, b_i, W^{\prime (c)}_i, b^{\prime (c)}$).

For the family of optimisation algorithms based on gradient descent,
we need to know the gradient of the likelihood
with respect to the model parameters,
which is given in~(\ref{eqn:diff}),
where $x\in \mathcal{X}$ is a latent entity,
and $c\in V$ is an observed predicate
(corresponding to the top and bottom rows of figure~\ref{fig:graph-token}).
Note that we extend the definition of energy
from situations to entities in the obvious way:
half the energy of an entity's links, plus its bias energy.
A full derivation of~(\ref{eqn:diff}) is given in the appendix.
\begin{equation}
\begin{split}
&\dth \log P(c) = \E{x|c}{\dthb{-E^b(x)}} \\
& - \E{x}{\dthb{-E^b(x)}} \\
& + \E{x|c}{\left(1-t_c(x)\right) \dthb{-E^p(x,c)}} \\
& - \E{x|c}{\E{\cp|x}{\left(1-t_\cp(x)\right) \dthb{-E^p(x,\cp)}}} \hspace*{-6mm}
\end{split}
\label{eqn:diff}
\end{equation}

There are four terms in this gradient:
the first two are for the background distribution,
and the last two are for the semantic functions.
In both cases, one term is positive, and conditioned on the data,
while the other term is negative, and represents the predictions of the model.

Calculating the expectations exactly is infeasible,
as this requires summing over all possible configurations.
Instead, we can use a Markov Chain Monte Carlo method,
as typically done for Latent Dirichlet Allocation \cite{blei2003lda,griffiths2004lda}.
Our aim is to sample values of $x$ and $c$,
and use these samples to approximate the expectations:
rather than summing over all values, we just consider the samples.
For each token in the training data,
we introduce a latent entity vector,
which we use to approximate the first, third, and fourth terms in~(\ref{eqn:diff}).
Additionally, we introduce a latent predicate for each latent entity,
which we use to approximate the fourth term --
this latent predicate is analogous to
the negative samples used by \newcite{mikolov2013vector}.

When resampling a latent entity conditioned on the data,
the conditional distribution $P(x|c)$ is unknown,
and calculating it directly requires summing over the whole semantic space.
For this reason, we cannot apply Gibbs sampling (as used in LDA),
which relies on knowing the conditional distribution.
However, if we compare two entities $x$ and $\xp$,
the normalisation constant cancels out in the ratio $P(\xp|c) / P(x|c)$,
so we can use the Metropolis-Hastings algorithm \cite{metropolis1953,hastings1970}.
Given the current sample~$x$,
we can uniformly choose one unit to switch on, and one to switch off,
to get a proposal~$\xp$.
If the ratio of probabilities shown in~(\ref{eqn:x-mh})
is above 1, we switch the sample to $\xp$;
if it's below 1, it is the probability of switching to $\xp$.
\begin{equation}
\frac{P(\xp|c)}{P(x|c)} = \frac{\ene{\xp}\frac{1}{Z_{\xp}}t_c(\xp)}{\ene{x}\frac{1}{Z_{x}}t_c(x)}
\label{eqn:x-mh}
\end{equation}

Although Metropolis-Hastings avoids the need to calculate
the normalisation constant $Z$ of the background distribution,
we still have the normalisation constant $Z_x$ of choosing a predicate given an entity.
This constant represents the number of predicates true of the entity
(weighted by frequency).
The intuitive explanation is that we should sample an entity
which few predicates are true of,
rather than an entity which many predicates are true of.
We approximate this constant by assuming that
we have an independent contribution from each dimension of $x$.
We first average over all predicates (weighted by frequency),
to get the average predicate~$W^{avg}$.
We then take the exponential of~$W^{avg}$
for the dimensions that we are proposing switching off and on --
intuitively, if many predicates have a large weight for a given dimension,
then many predicates will be true of an entity where that dimension is active.
This is shown in~(\ref{eqn:approx}),
where $x$ and $\xp$ differ in dimensions $i$ and $i^\prime$ only,
and where $k$ is a constant.
\begin{equation}
\frac{Z_x}{Z_\xp} \approx \exp\left(k \left(W^{avg}_i - W^{avg}_{i^\prime}\right)\right)
\label{eqn:approx}
\end{equation}

We must also resample latent predicates given a latent entity,
for the fourth term in~(\ref{eqn:diff}).
This can similarly be done using the Metropolis-Hastings algorithm,
according to the ratio shown in~(\ref{eqn:c-mh}).
\begin{equation}
\frac{P(\cp|x)}{P(c|x)} = \frac{f_\cp t_\cp(x)}{f_c t_c(x)}
\label{eqn:c-mh}
\end{equation}

Finally, we need to resample entities from the background distribution,
for the second term in~(\ref{eqn:diff}).
Rather than recalculating the samples from scratch after each weight update,
we used fantasy particles (persistent Markov chains),
which \newcite{tieleman2008particle} found effective for training RBMs.
Resampling a particle can be done more straightforwardly
than resampling the latent entities --
we can sample each entity conditioned on the other entities in the situation,
which can be done exactly using belief propagation
(see \newcite{yedidia2003belief} and references therein),
as \newcite{swersky2012cardinality} applied to the Cardinality RBM.

To make weight updates from the gradients,
we used AdaGrad \cite{duchi2011adagrad},
with exponential decay of the sum of squared gradients.
We also used L1 and L2 regularisation,
which determines our prior over model parameters.

We found that using a random initialisation is possible,
but seems to lead to a long training time, due to slow convergence.
We suspect that this could be because
the co-occurrence of predicates is mediated via at least two latent vectors,
which leads to mixing of semantic classes in each dimension, particularly in the early stages of training.
Such behaviour can happen with complicated topic models --
for example, \newcite{seaghdha2010latent} found this for their ``Dual Topic'' model.
One method to reduce convergence time
is to initialise predicate parameters using pre-trained vectors.
The link parameters can then be initialised as follows:
we consider a situation with just one entity,
and for each predicate, we find the mean-field entity vector
given the pre-trained predicate parameters;
we then fix all entity vectors in our training corpus to be these mean-field vectors,
and find the positive pointwise mutual information
of each each pair of entity dimensions, for each link label.
In particular, we initialised predicate parameters using our sparse SVO Word2Vec vectors,
which we describe in section~\ref{sec:eval}.

\section{Training and Initial Experiments}

In this section, we report the first experiments carried out within our framework.

\subsection{Training Data}
\label{sec:data}

Training our model requires a corpus of DMRS graphs.
In particular, we used WikiWoods,
an automatically parsed version of the July 2008 dump of the full English Wikipedia,
distributed by DELPH-IN\footnote{\url{http://moin.delph-in.net/WikiWoods}}.
This resource was produced by \newcite{flickinger2010wikiwoods},
using the English Resource Grammar (ERG; Flickinger, 2000),\nocite{flickinger2000erg}
trained on the manually treebanked subcorpus WeScience \cite{ytrestol2009wescience},
and implemented with the PET parser \cite{callmeier2001pet,toutanova2005parse}.
To preprocess the corpus, we used the python packages
pydelphin\footnote{\url{https://github.com/delph-in/pydelphin}} (developed by Michael Goodman),
and pydmrs\footnote{\url{https://github.com/delph-in/pydmrs}} \cite{copestake2016pydmrs}.

For simplicity, we restricted attention to subject-verb-object (SVO) triples,
although we should stress that this is not an inherent limitation of our model,
which could be applied to arbitrary graphs.
We searched for all verbs in the WikiWoods treebank,
excluding modals,
that had either an \textsc{arg1} or an \textsc{arg2}, or both.
We kept all instances whose arguments were nominal,
excluding pronouns and proper nouns.
The ERG does not automatically convert
out-of-vocabulary items from their surface form to lemmatised predicates,
so we applied WordNet's morphological processor Morphy \cite{fellbaum1998wordnet},
as available in NLTK \cite{bird2009nltk}.
Finally, we filtered out situations including rare predicates,
so that every predicate appears at least five times in the dataset.

As a result of this process, all data was of the form
(\textit{verb}, \textsc{arg1}, \textsc{arg2}),
where one (but not both) of the arguments may be missing.
A summary is given in table~\ref{tab:data}.
In total, the dataset contains 72m tokens,
with 88,526 distinct predicates.

\begin{table}[h]
\centering
\vspace*{-1mm}
\begin{tabular}{|l|r|}
\hline
Situation type & No.\ instances \\ \hline
Both arguments & 10,091,234 \\
\textsc{arg1} only & 6,301,280 \\
\textsc{arg2} only & 14,868,213 \\ \hline
Total & 31,260,727 \\ \hline
\end{tabular}
\vspace*{-1mm}
\caption{Size of the training data.}
\vspace*{-4mm}
\label{tab:data}
\end{table}

\subsection{Evaluation}
\label{sec:eval}

\begin{table*}
\center
\begin{tabular}{|l|c|c|c|c|c|}

\hline
Model & SimLex Nouns & SimLex Verbs & WordSim Sim. & WordSim Rel. \\ \hline
Word2Vec (10-word window) & .28 & .11 & \bf .69 & .46 \\

Word2Vec (2-word window) & .30 & .16 & .65 & .34 \\
SVO Word2Vec & .44 & .18 & .61 & .24 \\
Sparse SVO Word2Vec & \bf .45 & \bf .27 & .63 & .30 \\
Semantic Functions & .26 & .14 & .34 & \bf .01 \\ \hline

\end{tabular}

\caption{Spearman rank correlation of different models with average annotator judgements.
Note that we would like to have a \emph{low} score on the final column
(which measures relatedness, rather than similarity).
}
\label{tab:results}
\vspace*{-3mm}
\end{table*}

\begin{table}[t]
\centering
\begin{tabular}{|l|c|}
\hline
\emph{flood} / \emph{water} (related verb and noun) & .06 \\ 
\emph{flood} / \emph{water} (related nouns) & .43 \\
\emph{law} / \emph{lawyer} (related nouns) & .44 \\ \hline
\emph{sadness} / \emph{joy} (near-antonyms) & .77 \\
\emph{happiness} / \emph{joy} (near-synonyms) & .78 \\
\emph{aunt} / \emph{uncle} (differ in a single feature) & .90 \\
\emph{cat} / \emph{dog} (differ in many features) & .92 \\ \hline
\end{tabular}
\caption{Similarity scores for thematically related words,
and various types of co-hyponym.}
\label{tab:sim}
\vspace*{-3mm}
\end{table}

As our first attempt at evaluation,
we chose to look at two lexical similarity datasets.
The aim of this evaluation was simply to verify that the model was learning something reasonable.
We did not expect this task to illustrate our model's strengths,
since we need richer tasks to exploit its full expressiveness.
Both of our chosen datasets aim to evaluate similarity, rather than thematic relatedness:
the first is \newcite{hill2015simlex}'s SimLex-999 dataset,
and the second is \newcite{finkelstein2001wordsim}'s WordSim-353 dataset, 
which was split by \newcite{agirre2009wordsim}
into similarity and relatedness subsets.
So far, we have not tuned hyperparameters.

Results are given in table~\ref{tab:results}.
We also trained \newcite{mikolov2013vector}'s Word2Vec model
on the SVO data described in section~\ref{sec:data},
in order to give a direct comparison of models on the same training data.
In particular, we used the continuous bag-of-words model with negative sampling,
as implemented in \newcite{rehurek2010gensim}'s \emph{gensim} package,
with off-the-shelf hyperparameter settings.
We also converted these to sparse vectors
using \newcite{faruqui2015sparse}'s algorithm,
again using off-the-shelf hyperparameter settings.
To measure similarity of our semantic functions,
we treated each function's parameters as a vector
and used cosine similarity, for simplicity.

For comparison, we also include the performance of Word2Vec when trained on raw text.
For SimLex-999, we give the results reported by \newcite{hill2015simlex},
where the 2-word window model was the best performing model that they tested.
For WordSim-353, we trained a model on the full WikiWoods text,
after stripping all punctuation and converting to lowercase.
We used the \emph{gensim} implementation with off-the-shelf settings,
except for window size (2 or 10)
and dimension (200, as recommended by Hill et al.).
In fact, our re-trained model performed better on SimLex-999 than Hill et al.\ reported
(even when we used less preprocessing or a different edition of Wikipedia),
although still worse than our sparse SVO Word2Vec model.

It is interesting to note that training Word2Vec on verbs and their arguments
gives noticeably better results on SimLex-999 than training on full sentences,
even though far less data is being used:
$\sim$72m tokens, rather than $\sim$1000m.
The better performance suggests that
semantic dependencies may provide more informative contexts than simple word windows.
This is in line with previous results,
such as \newcite{levy2014dependency}'s work on using syntactic dependencies.
Nonetheless, this result deserves further investigation.

Of all the models we tested, only our semantic function model
failed on the relatedness subset of WordSim-353.
We take this as a positive result,
since it means the model clearly distinguishes
relatedness and similarity.

Examples of thematically related predicates and various kinds of co-hyponym
are given in table~\ref{tab:sim}, along with our model's similarity scores.
However, it is not clear that it is possible, or even desirable,
to represent these varied relationships on a single scale of similarity.
For example, it could be sensible to treat \emph{aunt} and \emph{uncle}
either as synonyms (they refer to relatives of the same degree of relatedness)
or as antonyms (they are ``opposite'' in some sense).
Which view is more appropriate will depend on the application, or on the context.

Nouns and verbs are very strongly distinguished,
which we would expect given the structure of our model.
This can be seen in the similarity scores between \emph{flood} and \emph{water},
when \emph{flood} is considered either as a verb or as a noun.\footnote{
  We considered the ERG predicates \texttt{\_flood\_v\_cause} and \texttt{\_flood\_n\_of},
  which were the most frequent predicates in WikiWoods for \emph{flood}, for each part of speech.}
SimLex-999 generally assigns low scores to near-antonyms,
and to pairs differing in a single feature,
which might explain why the performance of our model is not higher on this task.
However, the separation of thematically related predicates from co-hyponyms is a promising result.

\section{Related Work}

As mentioned above, \newcite{coecke2010tensor} and \newcite{baroni2014tensor}
introduce a tensor-based framework that incorporates argument structure through tensor contraction.
However, for logical inference,
we need to know how one vector can entail another.
\newcite{grefenstette2013logic} explores one method to do this;
however, they do not show that this approach is learnable from distributional information,
and furthermore, they prove that quantifiers cannot be expressed with tensors.


\newcite{balkir2014mixed}, working in the tensorial framework,
uses the quantum mechanical notion of a ``mixed state" to model uncertainty.
However, this doubles the number of tensor indices,
so squares the number of dimensions
(e.g.\ vectors become matrices).
In the original framework,
expressions with several arguments already have a high dimensionality
(e.g.\ \textit{whose} is represented by a fifth-order tensor),
and this problem becomes worse.

\newcite{vilnis2014gauss} embed predicates as Gaussian distributions over vectors.
By assuming covariances are diagonal, this only doubles the number of dimensions
($N$ dimensions for the mean, and $N$ for the covariances).
However, similarly to \newcite{mikolov2013vector},
they simply assume that nearby words have similar meanings,
so the model does not naturally capture compositionality or argument structure.

In both Balkır's and Vilnis and McCallum's models,
they use the probability of a vector given a word --
in the notation from section~\ref{sec:sem-func}, $P(x|t)$.
However, the opposite conditional probability, $P(t|x)$,
more easily allows composition.
For instance, if we know two predicates are true ($t_1$~and~$t_2$),
we cannot easily combine $P(x|t_1)$ and $P(x|t_2)$ to get $P(x|t_1,t_2)$ --
intuitively, we're generating $x$ twice.
In contrast, for semantic functions,
we can write $P(t_1,t_2|x)=P(t_1|x)P(t_2|x)$.

\newcite{gardenfors2004concept} argues concepts should be modelled as
convex subsets of a semantic space.
\newcite{erk2009region} builds on this idea,
but their model requires pre-trained count vectors,
while we learn our representations directly.
\newcite{mcmahan2015colour} also learn representations directly,
considering colour terms,
which are grounded in a well-understood perceptual space.
Instead of considering a single subset, they use
a probability distribution over subsets: $P(A|t)$ for $A\subset \mathcal{X}$.
This is more general than a semantic function $P(t|x)$,
since we can write ${P(t|x) = \sum_{A\ni v} P(A|t)}$.
However, this framework may be \emph{too} general,
since it means we cannot determine the truth of a predicate until we know the entire set $A$.
To avoid this issue, they factorise the distribution,
by assuming different boundaries of the set are independent.
However, this is equivalent to considering $P(t|x)$ directly,
along with some constraints on this function.
Indeed, for the experiments they describe, it is sufficient to know a semantic function $P(t|x)$.
Furthermore, McMahan and Stone find expressions like \textit{greenish}
which are nonconvex in perceptual space,
which suggests that representing concepts with convex sets
may not be the right way to go.

Our semantic functions are similar to
\newcite{cooper2015prob}'s probabilistic type judgements,
which they introduce within the framework of Type Theory with Records \cite{cooper2005ttr},
a rich semantic theory.
However, one difference between our models is that they represent situations in terms of situation types,
while we are careful to define our semantic space without reference to any predicates.
More practically,
although they outline how their model might be learned,
they assume we have access to type judgements for observed situations.
In contrast, we describe how a model can be learned from observed utterances,
which was necessary for us to train a model on a corpus.

\newcite{goodman2014prob} propose another linguistically motivated probabilistic model,
using the stochastic $\lambda$-calculus
(more concretely, probabilistic programs written in Church).
However, they rely on relatively complex generative processes,
specific to individual semantic domains,
where each word's meaning may be represented by a complex expression.
For a wide-scale system,
such structures would need to be extended to cover all concepts.
In contrast, our model assumes a direct mapping between predicates and semantic functions,
with a relatively simple generative structure determined by semantic dependencies.

Finally, our approach should be distinguished from
work which takes pre-trained distributional vectors,
and uses them within a richer semantic model.
For example,
\newcite{herbelot2015quantifier} construct a mapping
from a distributional vector
to judgements of which quantifier is most appropriate for a range of properties.
\newcite{erk2016alligator} uses distributional similarity
to probabilistically infer properties of one concept, given properties of another.
\newcite{beltagy2016combination} use distributional similarity
to produce weighted inference rules, which they incorporate in a Markov Logic Network.
Unlike these authors, we aim to directly learn interpretable representations,
rather than interpret given representations.

\section{Conclusion}

We have introduced a novel framework for distributional semantics,
where each predicate is represented as a function,
expressing how applicable the predicate is to different entities.
We have shown how this approach can capture semantic phenomena  
which are challenging for standard vector space models.
We have explained how our framework can be implemented,
and trained on a corpus of DMRS graphs.
Finally, our initial evaluation on similarity datasets demonstrates the feasibility of this approach,
and shows that thematically related words are not given similar representations.
In future work, we plan to use richer tasks which exploit the model's expressiveness.

\section*{Acknowledgments}

This work was funded by a Schiff Foundation Studentship.
We would also like to thank Yarin Gal,
who gave useful feedback on the specification of our generative model.


\bibliography{acl2016}
\bibliographystyle{acl2016}


\appendix

\section*{Appendix: Derivation of Gradients}
\label{sec:derivation}

In this section, we derive equation~(\ref{eqn:diff}).
As our model generates predicates from entities,
to find the probability of observing the predicates,
we need to sum over all possible entities.
After then applying the chain rule to log,
and expanding $P(x,c)$, we obtain the expression below.
\begin{equation*}
\begin{split}
&\dth \log P(c) = \dth \log \sum_x P(x,c) \\
&= \frac{\dth \sum_x P(x,c) }{\sum_\xp P(\xp,c)} \\
&= \frac{\dth \sum_x\frac{1}{Z_x} f_c t_c(x) \frac{1}{Z}\ene{x}}{\sum_\xp P(\xp,c)} 
\end{split}
\end{equation*}

\vspace{1mm}

When we now apply the product rule, we will get four terms,
but we can make use of the fact that the derivatives of all four terms
are multiples of the original term:
\begin{equation*}
\dth e^{-E^b(x)} = e^{-E^b(x)} \dthb{-E^b(x)}
\end{equation*}
\begin{equation*}
\dth t_c(x) = t_c(x)\left(1-t_c(x)\right) \dthb{-E^p(x,c)}
\end{equation*}
\begin{equation*}
\dth \frac{1}{Z_x} = \frac{-1}{Z_x^2}\dth Z_x
\end{equation*}
\begin{equation*}
\dth \frac{1}{Z} = \frac{-1}{Z^2}\dth Z
\end{equation*}

\vspace{3mm}

This allows us to derive:
\begin{equation*}
\begin{split}
=& \sum_x \frac{P(x,c)}{\sum_\xp P(\xp,c)} \left[ \dthb{-E^b(x)} \right. \\
&\hskip 15mm + \left(1-t_c(x)\right) \dthb{-E^p(x,c)} \\
&\hskip 15mm - \left. \frac{1}{Z_x}\dth Z_x \right] \\
&- \frac{\sum_x P(x,c)}{\sum_\xp P(\xp,c)}\frac{1}{Z}\dth Z
\end{split}
\end{equation*}


We can now simplify using conditional probabilities,
and expand the derivatives of the normalisation constants:
\begin{equation*}
\begin{split}
=& \sum_x P(x|c)\left[\dthb{-E^b(x)} \right. \\
&\hskip 20mm + \left(1-t_c(x)\right) \dthb{-E^p(x,c)} \\
&\hskip 20mm - \left. \frac{1}{Z_x}\dth \sum_\cp f_\cp t_\cp(x) \right] \\
&- \frac{1}{Z} \dth \sum_x\ene{x}
\end{split}
\end{equation*}
\begin{equation*}
\begin{split}
=& \sum_x P(x|c)\left[\dthb{-E^b(x)} \right. \\
&\hskip 3mm + \left(1-t_c(x)\right) \dthb{-E^p(x,c)} \\
&\hskip 3mm - \left. \sum_\cp\frac{f_\cp t_\cp(x)}{Z_x} \left(1-t_\cp(x)\right) \dthb{-E^p(x,\cp)} \right] \\
&- \sum_x \frac{\ene{x}}{Z} \dthb{-E^b(x)}
\end{split}
\end{equation*}
\begin{equation*}
\begin{split}
=& \sum_x P(x|c)\left[\dthb{-E^b(x)} \right. \\
&\hskip 3mm + \left(1-t_c(x)\right) \dthb{-E^p(x,c)} \\
&\hskip 3mm - \left. \sum_\cp P(\cp|x) \left(1-t_\cp(x)\right) \dthb{-E^p(x,\cp)} \right] \\
&- \sum_x P(x) \dthb{-E^b(x)}
\end{split}
\end{equation*}

\pagebreak

Finally, we write expectations instead of sums of probabilities:
\begin{equation*}
\begin{split}
=& \mathbb{E}_{x|c}\left[\dthb{-E^b(x)} \right. \\
&\hskip 8mm + \left(1-t_c(x)\right) \dthb{-E^p(x,c)} \\
&\hskip 8mm -\left. \E{\cp|x}{\left(1-t_\cp(x)\right) \dthb{-E^p(x,\cp)}} \right] \\
& - \E{x}{\dthb{-E^b(x)}} \\
\end{split}
\end{equation*}

\end{document}